\pdfoutput=1
\documentclass[10pt,twocolumn,letterpaper]{article}

\usepackage[pagenumbers]{cvpr}   

\usepackage{amsfonts}
\usepackage{bm}
\usepackage{float}
\usepackage{stfloats}
\usepackage{placeins}
\usepackage{multirow}
\usepackage{capt-of}
\makeatletter
\def\@dblfloatplacement{\global\@dbltopnum\c@dbltopnumber
   \global\@dbltoproom \dbltopfraction\@colht
   \@textmin \@colht \advance \@textmin -\@dbltoproom
   \@fpmin \dblfloatpagefraction\textheight
   \@fptop \@dblfptop
   \@fpsep \@dblfpsep
   \@fpbot \@dblfpbot
   \def\f@depth{1sp}}%
\makeatother
\usepackage{microtype}

\def\1{\bm{1}}

\DeclareMathAlphabet{\mathsfit}{\encodingdefault}{\sfdefault}{m}{sl}
\SetMathAlphabet{\mathsfit}{bold}{\encodingdefault}{\sfdefault}{bx}{n}

\makeatletter
\renewcommand{\paragraph}{\@startsection{paragraph}{4}{\z@}{5pt plus 1.5pt minus 1pt}{-1em}{\normalsize\bfseries}}
\makeatother

\usepackage{tcolorbox}
\usepackage{datetime}
\newdateformat{monthyeardate}{\monthname[\THEMONTH] \THEYEAR}

\definecolor{cvprblue}{rgb}{0.21,0.49,0.74}
\usepackage[breaklinks,colorlinks,allcolors=cvprblue]{hyperref}

\newcommand{\papertitlemultiline}{OpenFlyV2: An onboard vision-language-action model\\for aerial perception and navigation}
\newcommand{\projectpageurl}{https://shailab-ipec.github.io/OpenFlyV2/}
\newcommand{\githuburl}{https://github.com/SHAILAB-IPEC/OpenFlyV2}

\def\confName{CVPR}
\def\confYear{2027}

\title{\papertitlemultiline}
\author{}

\definecolor{openflynavy}{HTML}{0A1640}
\definecolor{pillbg}{HTML}{F5F8FE}
\definecolor{pillborder}{HTML}{CBD8EE}
\definecolor{affilgray}{HTML}{4A5568}
\definecolor{ruleblue}{HTML}{CBD6E8}

\def\keye{\raisebox{-1.8pt}{\includegraphics[height=1.02em]{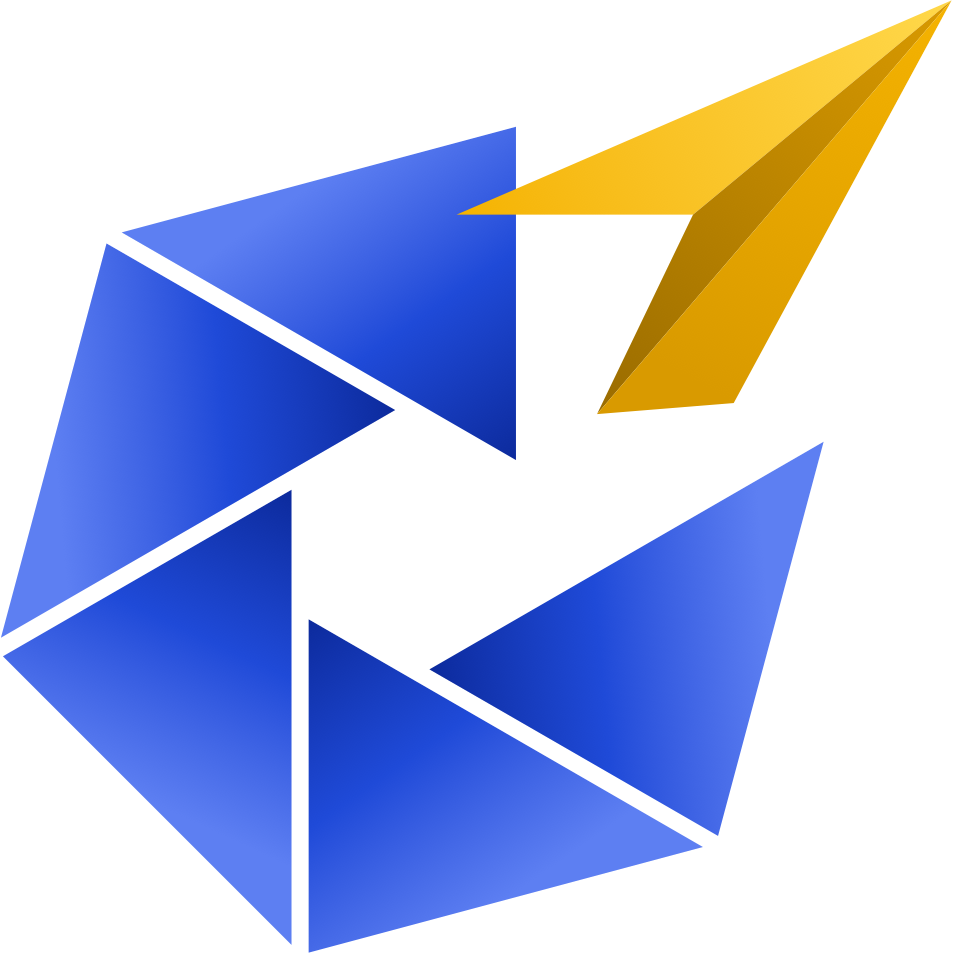}}}
\def\github{\raisebox{-1.8pt}{\includegraphics[height=1.02em]{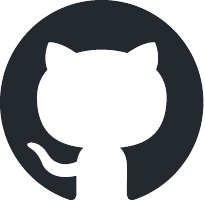}}}

\newcommand{\linkpill}[3]{%
  \href{#2}{#1\hspace{3.5pt}{\fontsize{9}{11}\selectfont\color{cvprblue}#3}}%
}

\usepackage[T1]{fontenc}
\usepackage{array,longtable}
\newcolumntype{L}[1]{>{\raggedright\arraybackslash}p{#1}}
\newcolumntype{R}[1]{>{\raggedleft\arraybackslash}p{#1}}
\newcolumntype{C}[1]{>{\centering\arraybackslash}p{#1}}

\begin{document}

\twocolumn[{%
\renewcommand\twocolumn[1][]{#1}%
\vspace*{-26pt}%
\noindent
\begin{minipage}[b]{0.55\textwidth}
  \includegraphics[height=18pt]{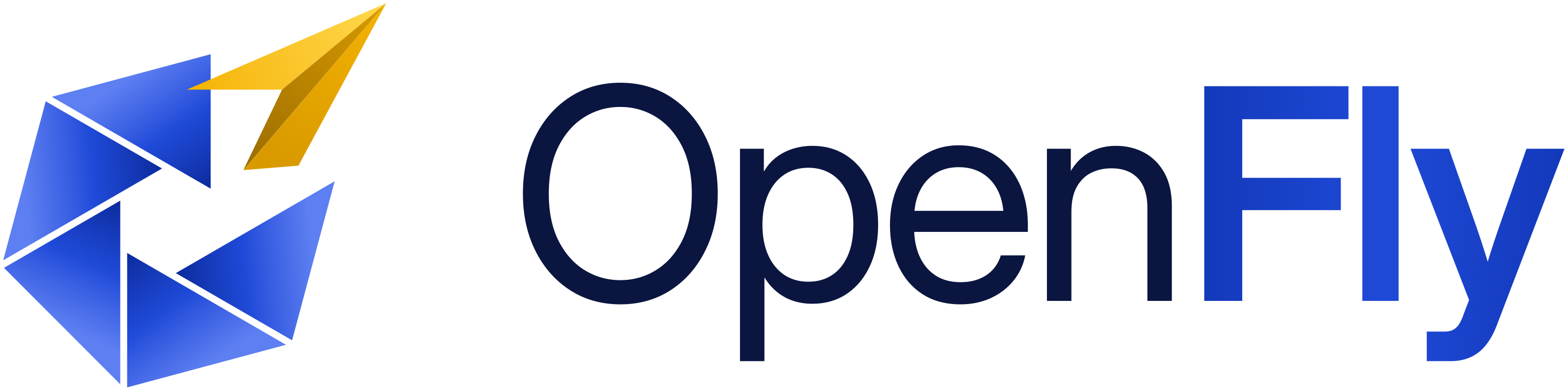}%
\end{minipage}%
\hfill
\begin{minipage}[b]{0.42\textwidth}
  \raggedleft
  {\fontsize{8}{10}\selectfont\itshape\color{affilgray}\monthyeardate\today}%
\end{minipage}\par
\vspace{4pt}%
\noindent{\color{ruleblue}\hrule height 0.7pt}\par
\vspace{12pt}%
\begin{center}
  {\fontsize{15.5}{19.0}\selectfont\bfseries\color{openflynavy}%
    \papertitlemultiline\par}%
  \vspace{7pt}%
  {\fontsize{10}{13}\selectfont

    Junli~Liu$^{1,2}$ \quad Zhongrui~You$^{1,3}$ \quad Chenhui~Li$^{1}$ \quad Zhen~Li$^{1,4}$ \quad Yuyang~Tian$^{1,5}$ \quad Qizhi~Chen$^{1,6}$\\
    Delin~Qu$^{1}$ \quad Haoming~Song$^{1,4}$ \quad Dong~Wang$^{1}$ \quad Zhigang~Wang$^{1}$ \quad Bin~Zhao$^{1,2}$
    \par}
  \vspace{4pt}%
  {\fontsize{8.5}{10.5}\selectfont\color{affilgray}
    $^{1}$Shanghai Artificial Intelligence Laboratory\enspace $^{2}$Northwestern Polytechnical University\enspace $^{3}$Beihang University\\
    $^{4}$Shanghai Jiao Tong University\enspace $^{5}$University of Science and Technology of China\enspace $^{6}$Zhejiang University\par}
  \vspace{7pt}%
  \linkpill{\keye}{\projectpageurl}{Project Page}\hspace{7pt}%
  \linkpill{\github}{\githuburl}{GitHub}\hspace{7pt}%

\par\vspace{6pt}%
    \centering
    \captionsetup{type=figure}
    \includegraphics[width=0.99\textwidth]{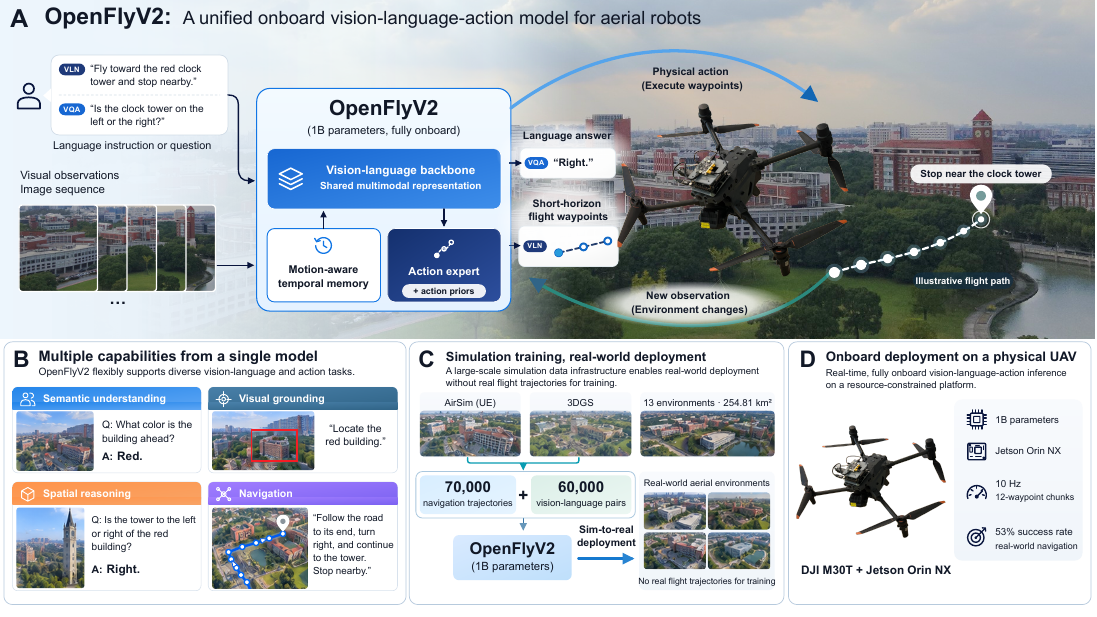}\vspace{-3pt}
    \captionof{figure}{\textbf{Overview of OpenFlyV2, a unified onboard vision-language-action model for aerial robots.} \textbf{(A)} A shared vision-language backbone integrates visual observations and language instructions with motion-aware temporal memory (Motion-Adaptive Memory, MAM); an action expert with action priors (Motion-Structured Prior, MSP) generates short-horizon waypoints for closed-loop flight. \textbf{(B)} A single model supports semantic understanding, visual grounding, spatial reasoning, and navigation. \textbf{(C)} OpenFlyV2-130K comprises approximately 70,000 navigation trajectories and 60,000 vision-language pairs from 13 environments spanning 254.81~km\textsuperscript{2}. \textbf{(D)} The 1-billion-parameter model generates 12-waypoint action chunks at 10~Hz entirely onboard a DJI M30T equipped with an NVIDIA Jetson Orin NX, achieving a 53\% success rate in real-world navigation without training on real-world flight trajectories.}
    \label{fig:1}
\end{center}%
\vspace{2pt}%
}]

\begin{abstract}
Human cognition flexibly draws on a shared perceptual foundation to understand environments, attend to referenced entities, reason about spatial relationships, and act. Endowing aerial robots with an analogous functional organization remains challenging, as existing systems are often specialized for individual capabilities, dependent on offboard computation, or unable to sustain closed-loop operation on resource-constrained platforms. We present OpenFlyV2, a unified onboard vision-language-action architecture that uses a shared vision-language representation to support semantic understanding, language-grounded perception, spatial reasoning, and continuous aerial control within a single model. A compact action expert converts the shared multimodal context into short-horizon flight waypoints, while motion-aware temporal memory and action priors reduce the computational cost of closed-loop execution. We construct OpenFlyV2-130K, a simulation-based dataset comprising approximately 70,000 navigation trajectories and 60,000 vision-language instruction--response pairs. OpenFlyV2 generates action chunks, each comprising 12 waypoints, entirely onboard at 10 hertz using an NVIDIA Jetson Orin NX. It transfers from simulation to physical quadrotor flight without training on real-world flight trajectories, achieving a 53.33\% navigation success rate in 300 trials across 10 real-world environments. These results show that broad vision-language competence and continuous closed-loop flight can be unified within a single model running on a resource-constrained aerial platform.
\end{abstract}

\section{Introduction}

Human intelligence is characterized by the ability to flexibly coordinate perception, reasoning, and action according to changing goals~\cite{r1}. Rather than treating each capability as an isolated task, different functions can be invoked as needed and coordinated within a common cognitive system~\cite{r2,r3}. This functional flexibility suggests an important goal for embodied intelligence: enabling a single system to support diverse capabilities and engage them according to task demands.

For aerial robots, this functional flexibility requires more than translating language instructions into flight trajectories: an aerial system may also need to interpret scene semantics, ground language-referenced entities, and reason about spatial relationships according to task demands. These capabilities have largely been studied separately. Vision-language navigation (VLN) connects linguistic goals with visual observations and action, from instruction-following benchmarks on navigation graphs and in continuous environments~\cite{r4,r5} to vision-language-action models deployed on legged robots~\cite{r6}, whereas visual question answering (VQA) and referring-expression grounding for robots examine complementary aspects of understanding and language grounding~\cite{r7,r8}. Aerial research has followed the navigation line, progressing from mapping navigation instructions to quadcopter control in simulation~\cite{r9} to datasets and platforms for landmark-directed navigation~\cite{r10,r11} and models for direct language-conditioned control~\cite{r12}, leaving scene understanding and spatial reasoning to separate systems.

Evidence from ground robots suggests that these capabilities benefit from being learned together. Joint training on vision-language and embodied data outperformed training on embodied data alone~\cite{r13}, and co-fine-tuning on web-scale vision-language data enabled a robot policy to follow instructions involving objects and concepts absent from its robot data~\cite{r14}. Yet these gains were obtained with models of billions of parameters running offboard. A central challenge is therefore to bring these capabilities together in a single aerial model that can also sustain closed-loop navigation entirely onboard. Offboard inference can alleviate local computational constraints, but makes the navigation loop dependent on communication availability and latency. Onboard execution removes this dependence, and onboard systems have achieved high-speed flight and swarm navigation in the wild with all computation on the aircraft~\cite{r15,r16}. These systems, however, address geometric tasks such as obstacle avoidance and coordinated flight and do not take language instructions as goals. Extending onboard execution to language-conditioned navigation requires multimodal understanding, temporal context processing, and waypoint generation, all within limited computational resources and fast enough to respond to new observations during flight. Together, these requirements call for a unified aerial vision-language-action model that combines scene understanding and language-conditioned flight with the computational efficiency needed for closed-loop operation onboard.

Building a unified onboard model poses three coupled challenges in capability integration, supervision, and onboard execution. First, a single model must support diverse vision-language capabilities while also connecting the same perceptual and linguistic context to continuous flight behavior when action is required. Second, training requires large-scale aerial supervision that links scene semantics and spatial relationships to language-conditioned flight behavior across changing viewpoints and diverse three-dimensional environments. Third, the model must process visual history and generate continuous waypoint sequences while meeting the latency constraints of onboard hardware. This constraint is especially important for aerial navigation: an unmanned aerial vehicle (UAV) that must slow down or hover while awaiting each prediction loses average flight speed and mission efficiency and cannot exploit its inherent agility~\cite{r17,r18}. These challenges make capability integration, scalable supervision, and computational efficiency inseparable considerations in the design of an onboard aerial vision-language-action system.

To address these challenges, we developed OpenFlyV2, an onboard vision-language-action model that integrates vision-language understanding and flight control within a single architecture. A shared vision-language backbone processes the current observation, language input, and historical context, and supports semantic understanding, visual grounding, and spatial reasoning. When physical action is required, a compact action expert operates on the same multimodal context to generate short-horizon flight waypoints. The individual capabilities can be accessed according to task demands.

Within this architecture, two components address the computational demands of onboard execution. Motion-Adaptive Memory (MAM) compresses flight history into a bounded set of event-level tokens by combining motion segments with their associated visual anchors. Temporal context allows past observations and executed movements to inform instruction interpretation~\cite{r19}, but representing it through topological graphs or dense video histories~\cite{r20,r21,r22} increases the computational cost of each decision; MAM retains motion-conditioned visual context while limiting the historical input to the vision-language backbone. The Motion-Structured Prior (MSP) exploits the observation that short-horizon UAV trajectories are concentrated around a small set of recurring motion patterns. It constructs a context-conditioned source distribution around a selected and adjusted motion prototype for flow-based generation, enabling the action expert to generate waypoint sequences with fewer flow-integration steps. Together, MAM and MSP reduce the costs of processing visual history and generating waypoint sequences, enabling efficient closed-loop navigation on resource-constrained onboard hardware.

Training OpenFlyV2 required supervision spanning both vision-language understanding and aerial navigation. We therefore constructed a scalable data-generation pipeline drawing on AirSim environments~\cite{r23}, Unreal Engine (UE) scenes, and real-world scenes reconstructed by our team using 3D Gaussian Splatting (3DGS)~\cite{r24}. This pipeline generated approximately 70,000 language-conditioned navigation trajectories and 60,000 aerial perception-and-reasoning instruction--response pairs across 13 environments. A subset of the navigation trajectories provides supervision for waypoint prediction, whereas the instruction--response pairs cover semantic understanding, visual grounding, and spatial reasoning, enabling these capabilities to be learned jointly with navigation. Language annotations were generated automatically and subsequently verified by human annotators.

We evaluated OpenFlyV2 across vision-language capabilities, simulated navigation, onboard inference, and physical flight. A single 1-billion-parameter checkpoint matches or exceeds three general-purpose vision-language models, including a 9-billion-parameter open model and two proprietary models, in 14 of 15 model--task comparisons spanning semantic understanding, visual grounding, and spatial reasoning. In simulated language-conditioned navigation, it outperforms four open-source VLN methods, exceeding the success rate (SR) of the strongest baseline by 24.4 and 14.7 percentage points in seen and unseen environments, respectively. Ablation studies quantify the contributions of MAM and MSP to navigation performance and computational efficiency. On an NVIDIA Jetson Orin NX, OpenFlyV2 generates action chunks at 10 Hz, supporting closed-loop navigation entirely onboard. We further deployed the same checkpoint on a physical quadrotor in 10 real-world environments, including university campuses, factories, parks, and urban areas, where it achieved a navigation SR of 53.33\% in 300 trials without training on real-world flight trajectories. These results show that broad aerial vision-language capabilities and closed-loop flight can be integrated within a single model operating on a resource-constrained aerial platform.

\section{Results}

\subsection{System overview}

OpenFlyV2 comprises a Qwen3.5-0.8B~\cite{r25} backbone and a 0.2-billion-parameter action expert (Fig.~1). The backbone processes aerial observations and language input to produce textual responses or condition waypoint generation. We used one final checkpoint for the main vision-language, simulated navigation, and physical flight evaluations reported below.

During navigation, the model uses current and historical observations to predict waypoints that an onboard path tracker converts into velocity commands for the flight controller. Two components adapt this loop to limited onboard resources: MAM retains visual-action history in compact event tokens, and MSP initializes action generation from a context-conditioned motion distribution. Supervised-model ablations assess their contributions to navigation before reinforcement post-training.

\subsection{Shared environments for aerial navigation and visual question answering}

We constructed OpenFlyV2-130K to support VLN and VQA within the same collection of 13 environments (Fig.~2A). Four AirSim environments and two additional UE city environments provided engine-authored scenes; seven environments reconstructed using 3DGS, comprising five university campuses and two post-disaster ruins, contributed layouts and appearance derived from real imagery. Data from 10 environments were used for training; data from the remaining three (AirSim 16, UE Big City, and GS NWPU 01) were excluded from both VLN and VQA training.

\begin{figure*}[t]
\centering
\includegraphics[width=\textwidth]{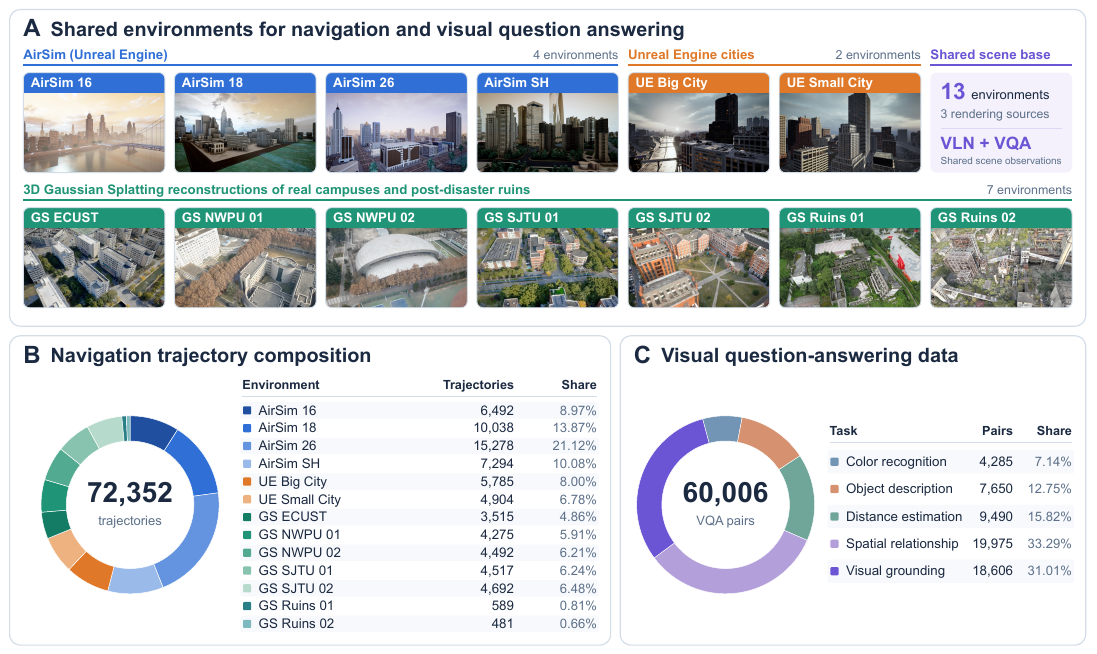}
\caption{\textbf{Shared environments and data composition of OpenFlyV2-130K.} \textbf{(A)} Thirteen environments supporting both navigation and VQA: four AirSim environments, two Unreal Engine city environments, and seven 3DGS reconstructions of real campuses and post-disaster ruins. \textbf{(B)} Distribution of 72,352 navigation trajectories across the environments. \textbf{(C)} Distribution of 60,006 VQA pairs across color recognition, object description, distance estimation, spatial relationship reasoning, and visual grounding.}
\label{fig:2}
\end{figure*}

The navigation collection comprises 72,352 trajectories (Fig.~2B), pairing observation and action sequences with language instructions. The VQA collection comprises 60,006 question--answer pairs covering color recognition, object description, distance estimation, spatial relationship reasoning, and visual grounding (Fig.~2C). Spatial questions probe the relationship between two objects along complementary dimensions: relative position in the image plane, relative distance with respect to the camera, topological relations such as overlap and containment, and relative size. They range from single pairwise comparisons to compositional descriptions that integrate multiple relations. Navigation thus supplies sequential action supervision, whereas VQA supplies questions and answers about individual aerial views, with responses expressed in language or image coordinates. Dataset preparation and the training subset are detailed in section S1 and table S1.

\subsection{Aerial vision-language understanding and spatial reasoning}

With a 0.8-billion-parameter vision-language backbone, OpenFlyV2 recognizes object attributes, describes aerial scenes, estimates distances, reasons about spatial relationships, and grounds language-referenced targets in the image (Fig. 3). It matched or exceeded the 9-billion-parameter Qwen3.5-9B~\cite{r26} and Gemini 3 Flash~\cite{r27} on all five vision-language tasks and GPT-5.5~\cite{r28} on four of the five (Fig. 4A). These capabilities allow the same model that supports onboard navigation to interpret what it sees and answer questions about the surrounding environment.

\begin{figure*}[t]
\centering
\includegraphics[width=\textwidth]{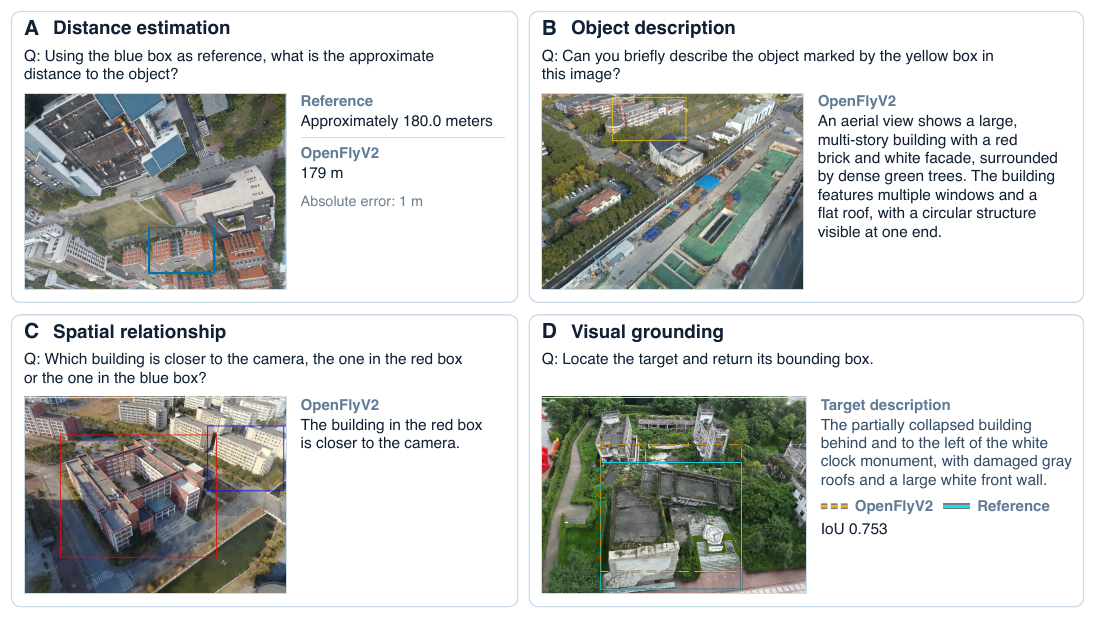}
\caption{\textbf{Aerial vision-language understanding and spatial reasoning with OpenFlyV2.} \textbf{(A)} Distance estimation for the building marked in blue: 179~m compared with a reference distance of 180~m. \textbf{(B)} Description of the building within the yellow box. \textbf{(C)} Relative-distance reasoning identifies the red-box building as closer to the camera than the blue-box building. \textbf{(D)} Localization of a language-described target. The predicted bounding box (dashed orange) and reference box (solid cyan) have an intersection over union (IoU) of 0.753.}
\label{fig:3}
\end{figure*}

\begin{figure*}[t]
\centering
\includegraphics[width=\textwidth]{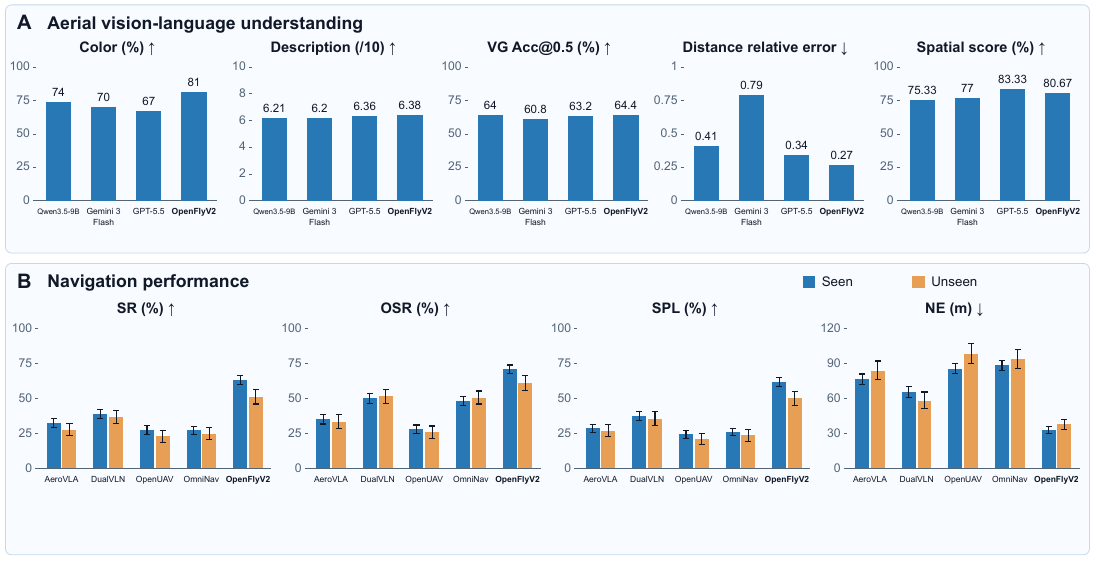}
\caption{\textbf{Quantitative evaluation of aerial understanding and navigation.} \textbf{(A)} Comparison with general-purpose vision-language models on color recognition, object description, visual grounding, distance estimation, and spatial relationship reasoning (300 examples per task; n = 250 visible-target examples for visual grounding). Visual grounding is reported as accuracy at an IoU threshold of 0.5 (Acc@0.5), and distance estimation is shown as relative error; values with 95\% bootstrap confidence intervals are listed in table S8. \textbf{(B)} Navigation performance in seen and unseen simulated environments, measured by success rate (SR), oracle success rate (OSR), success weighted by path length (SPL), and navigation error (NE). The navigation baselines were trained on our aerial navigation dataset and evaluated on the same 800 seen and 300 unseen episodes as OpenFlyV2. Success requires a terminal goal distance below 25 m and no detected collision throughout the episode. Blue and orange bars in (B) indicate seen and unseen environments, respectively. Arrows indicate the direction of better performance. Error bars in (B) show 95\% confidence intervals from 10,000 bootstrap resamples stratified by environment. \textit{P} values for paired comparisons with OpenFlyV2 are listed in table S11.}
\label{fig:4}
\end{figure*}

OpenFlyV2 achieved the highest color-recognition accuracy among the compared models (81.00\%; adjusted \textit{P} $\leq$ 0.042 against each). Its visual-grounding accuracy at an intersection over union (IoU) threshold of 0.5 (Acc@0.5; 64.40\%) was on par with those of Qwen3.5-9B (64.00\%) and GPT-5.5 (63.20\%), which differed by one and three of 250 examples, respectively. OpenFlyV2 also had a lower relative distance-estimation error than all three models (0.27 versus 0.34 for GPT-5.5; adjusted \textit{P} < 0.001 for each), whereas its mean absolute error did not differ significantly from that of GPT-5.5 (47.32 versus 52.62 m; adjusted \textit{P} = 0.11). The mean description scores of OpenFlyV2 and GPT-5.5 were similar (6.38 versus 6.36 out of 10), while GPT-5.5 had higher accuracy on spatial relationship reasoning (83.33\% versus 80.67\%). Differences in visual grounding, description score, and spatial relationship reasoning between OpenFlyV2 and each general-purpose model were not significant (adjusted \textit{P} $\geq$ 0.13). Thus, the 0.8-billion-parameter backbone that drives onboard navigation, about one-eleventh the size of Qwen3.5-9B, provided aerial perception comparable to or better than that of all evaluated general-purpose models.

\subsection{Vision-language navigation in simulated environments}

Navigation was evaluated on 800 episodes in seen environments and 300 in unseen environments (Fig.~4B). Four open-source baselines---AeroVLA~\cite{r12}, DualVLN~\cite{r29}, the navigation model from OpenUAV~\cite{r10}, and OmniNav~\cite{r30}---retained their original designs and training procedures, using only our navigation data for task-specific training. Because of their model size, none of these baselines fit within the 16 GB of memory on the NVIDIA Jetson Orin NX used for physical flight, so they were compared with OpenFlyV2 in simulation only. All methods were evaluated on the same episodes, with success requiring a terminal goal distance below 25 m and no detected collision throughout the episode.

OpenFlyV2 outperformed the other VLN models in both seen and unseen environments, achieving SR of 63.25\% (506 of 800 episodes) and 51.33\% (154 of 300), respectively. Collisions occurred in 2.8\% of its simulated episodes (31 of 1,100). These SRs exceeded the corresponding results of DualVLN (38.88\% and 36.67\%) by 24.38 and 14.67 percentage points. This advantage did not depend on reinforcement post-training: before post-training, the supervised MAM + MSP checkpoint reached a pooled SR of 57.09\% across all 1,100 episodes (Table 1), compared with 38.27\% (421 of 1,100) for DualVLN.

\begin{table*}[t]
\centering
\caption{ \textbf{Navigation ablations.} All variants underwent supervised training. Metrics are pooled across the 800 seen and 300 unseen evaluation episodes. The Steps column reports the number of flow-integration steps; SR, OSR, and SPL are percentages, and NE is in meters. Arrows indicate the direction of better performance. Bold values mark the best result in each metric across all six configurations. Values in brackets are 95\% Wilson score confidence intervals for SR and OSR (n = 1,100 episodes).}\label{tab:1}
\small
\setlength{\tabcolsep}{8pt}
\begin{tabular}{lccccc}
\toprule
\textbf{Model} & \textbf{Steps} & \textbf{SR (\%) $\uparrow$} & \textbf{OSR (\%) $\uparrow$} & \textbf{SPL (\%) $\uparrow$} & \textbf{NE (m) $\downarrow$} \\
\midrule
Base & 1 & 49.82~{\scriptsize [46.87, 52.77]} & 54.27~{\scriptsize [51.32, 57.20]} & 45.59 & 49.18 \\
Base & 2 & 56.36~{\scriptsize [53.42, 59.27]} & 60.64~{\scriptsize [57.72, 63.48]} & 53.32 & 42.35 \\
MAM & 1 & 53.45~{\scriptsize [50.50, 56.39]} & 63.18~{\scriptsize [60.29, 65.98]} & 52.22 & 37.54 \\
MAM & 2 & \textbf{59.45}~{\scriptsize [56.53, 62.32]} & \textbf{69.45}~{\scriptsize [66.67, 72.10]} & \textbf{58.02} & \textbf{32.65} \\
MAM + MSP & 1 & 57.09~{\scriptsize [54.15, 59.99]} & 68.36~{\scriptsize [65.56, 71.04]} & 55.82 & 33.55 \\
MAM + MSP & 2 & 57.73~{\scriptsize [54.79, 60.61]} & 68.45~{\scriptsize [65.65, 71.13]} & 56.36 & 34.57 \\
\bottomrule
\end{tabular}
\end{table*}

The higher completion rates were accompanied by lower terminal navigation error (NE). In seen environments, NE was 32.87 m for OpenFlyV2 and 65.75 m for DualVLN; in unseen environments, the corresponding errors were 37.73 and 57.98 m.

OpenFlyV2 also obtained the highest oracle success rate (OSR) and success weighted by path length (SPL) in both sets. Together, these measures show improvements in task completion, goal reaching, and success-weighted path efficiency. The differences from each baseline were significant for all four metrics in both sets (adjusted \textit{P} $\leq$ 0.030).

\subsection{History compression and single-step action generation}

We compared Base, MAM, and MAM + MSP at one and two flow-integration steps (Table~1). Metrics were pooled across the 800 seen and 300 unseen episodes. The main navigation comparison and physical deployment used the final post-trained policy with one integration step per action chunk. Model implementation, training, and evaluation details are given in sections S2 to S4, with the supervised schedule, terminal rewards, and rollout configuration in tables S2 to S4.

The Base configuration, with four historical images and the current image, used 446 input tokens. Replacing the historical-image inputs with MAM event tokens reduced this count to at most 218, a reduction of at least 51.1\%, while representing a longer flight history. Both counts include the instruction and action tokens in addition to the visual or memory tokens. With both configurations using one flow-integration step, action-expert inference on the NVIDIA Jetson Orin NX took 80~ms for Base and 20~ms with MAM, a fourfold speedup in this stage. The measurement configuration is detailed in section S5.

MAM also improved all four navigation metrics at both integration budgets. With one step, SR increased from 49.82\% to 53.45\%, and NE decreased from 49.18 to 37.54 m. With two steps, SR increased from 56.36\% to 59.45\%, with corresponding improvements in OSR and SPL.

Under single-step generation, adding MSP to MAM raised SR from 53.45\% to 57.09\% and reduced NE from 37.54 to 33.55 m. The combined one-step model also scored higher than the two-step Base on all four metrics, although the SR margin was small (57.09\% versus 56.36\%).

The benefit of MSP was concentrated at the single-step budget used onboard, where it recovered 61\% of the SR gain that MAM obtained from a second integration step, without the additional integration pass. Consistent with this role, a second step changed the SR of the combined model only from 57.09\% to 57.73\%, whereas Base SR increased from 49.82\% to 56.36\%, suggesting that the motion-structured source already supplies much of the refinement that a second step otherwise provides. With two steps, MAM alone had the highest SR (59.45\% versus 57.73\% for the combined model).

\subsection{Real-world onboard aerial vision-language navigation}

For physical flight, we deployed OpenFlyV2 on a DJI M30T equipped with an NVIDIA Jetson Orin NX, without additional training on real-world flight trajectories or test-site fine-tuning. Evaluation covered 10 real-world environments spanning university campuses, factories, parks, and urban areas. Of these, 2 were represented among the 3DGS training environments, and the remaining 8 were unseen locations.

The 300 trials comprised 150 single-landmark and 150 two-landmark instructions. A human observer judged each outcome: success required the flight path to follow the instruction and the specified destination landmark to appear in the central or lower-central region of the final camera image, and flights that clearly deviated from the instructed route were terminated and counted as failures.

Across these trials, OpenFlyV2 achieved an aggregate SR of 53.33\% (160 of 300 trials; 95\% confidence interval, 47.7 to 58.9\%). The SR was 54.51\% (127 of 233 trials) at the unseen sites and 49.25\% (33 of 67) at the sites represented in training. The three campus sites, including both seen sites, had an SR of 45.88\% (39 of 85 trials), whereas the park, factory, and urban sites, all unseen, had an SR of 56.28\% (121 of 215), with similar proportions of single-landmark instructions (48\% and 51\%). A likely reason is that campus buildings often share a uniform architectural style, which makes landmark descriptions less distinctive, whereas buildings at the other sites more often have individual styles that are easier to identify. Successful flights included sequential landmark-directed goals and instruction-dependent turns (Fig.~5), requiring repeated interpretation of the language goal as motion changed the available observations. Results by site and landmark count are reported in tables S5 and S6 (section S5). Selected successful two-landmark flights, one from each of four sites, are shown in movie S1. Three additional successful approaches to the same target from different initial poses are shown in fig. S1 and movie S2. Selected unsuccessful trials, one from each of four sites, are shown in movie S3.

\begin{figure*}[t]
\centering
\includegraphics[width=\textwidth]{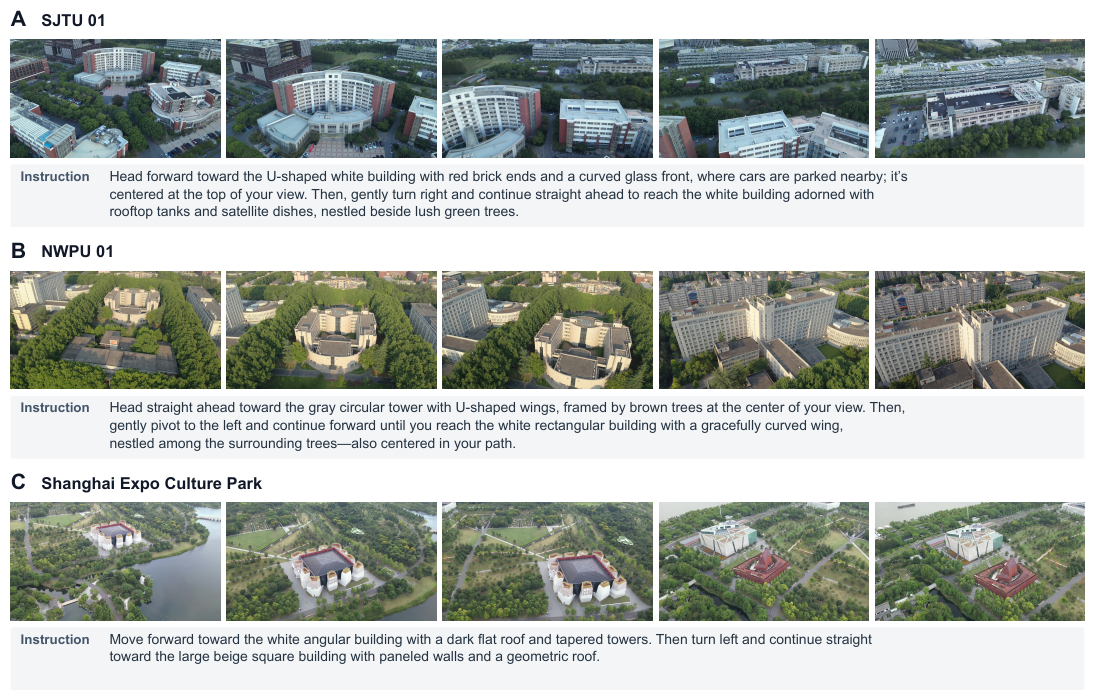}
\caption{\textbf{Onboard aerial vision-language navigation.} Three physical flight trials performed by OpenFlyV2 on a DJI M30T, with all navigation inference executed on an onboard NVIDIA Jetson Orin NX. \textbf{(A)} SJTU 01, a seen site. \textbf{(B)} NWPU 01, an unseen site. \textbf{(C)} Shanghai Expo Culture Park, an unseen site. Each row shows a successful two-landmark navigation trial, with successive camera observations arranged from left to right. The instructions require approaching the first landmark, turning, and continuing toward the second. Seen sites were represented in the 3DGS training environments.}
\label{fig:5}
\end{figure*}

During physical flight, the final policy used one flow-integration step to generate each 12-waypoint action chunk at 10~Hz on the Jetson Orin NX, without weight quantization, pruning, or distillation. Visual processing, memory updates, and action generation ran on the quadrotor. Asynchronous execution overlapped generation of a new chunk with execution of the preceding one, while an onboard path tracker generated body-frame velocity and yaw-rate commands and the flight controller handled stabilization. These trials demonstrate transfer from rendered observations to fully onboard physical navigation. The onboard efficiency measurements are summarized in table S7.

\section{Discussion}

Earlier vision-language-action models established that vision-language pretraining can support robotic control~\cite{r14,r31}. OpenFlyV2 brings scene understanding and language-conditioned navigation into an aerial system with a shared representation and a compact action expert. With a 0.8-billion-parameter vision-language backbone, it matches or exceeds the compared general-purpose vision-language models in 14 of 15 model--task comparisons and outperforms all four tested navigation methods in simulation. Because both functions run within the quadrotor's onboard computational budget, an operator can query the scene and direct the flight in natural language, while the perception--action loop remains independent of offboard computation and communication latency.

The ablations indicate that the representation of past observations and the initialization of future motion address distinct resource constraints. MAM preserves useful history through motion-conditioned compression, reducing the input token count and action-expert inference time while improving navigation. MSP supplies motion structure before refinement begins, improving single-step generation. This single-step improvement makes structured initialization useful for waypoint generation under onboard computational constraints. More broadly, these results show that design choices made for onboard efficiency need not trade off against task performance: MAM reduced both computational cost and navigation error, and MSP recovered more than half of the SR gain that MAM obtained from a second flow-integration step, without the additional integration pass.

Previous work has demonstrated the transfer of simulation-trained visuomotor policies to physical flight~\cite{r15,r32}, with domain randomization reducing the visual simulation-to-reality gap in drone racing~\cite{r33}. Transfer has also been studied for real-world object navigation on ground robots~\cite{r34}. Our physical trials bring language-conditioned landmark interpretation and task progression into a fully onboard perception--action loop across 10 environments spanning university campuses, factories, parks, and urban areas. The SR at the 8 sites absent from the training data was no lower than that at the 2 sites represented in training, indicating that landmark interpretation learned from rendered observations transferred beyond the reconstructed scenes. Coupling language-conditioned onboard policies with safety-assured high-speed navigation~\cite{r35} and with flight policies that remain robust out of distribution~\cite{r36} could extend these capabilities to faster and safety-critical missions.

This study has several limitations, each of which suggests a direction for future work. First, the 140 failed physical trials fell mainly into two modes: the model misidentified the target landmark or turned excessively and departed from the instructed route (movie S3). Together with the lower SR at campus sites, where buildings share a uniform architectural style, the first mode suggests that physical performance is limited more by distinguishing visually similar landmarks than by generating motion. The second suggests that short-horizon action chunks lack a mechanism for recognizing accumulated heading error. Explicit landmark verification before committing to a route, and recovery behavior learned from closed-loop experience, are therefore natural next steps. Second, MSP relies on four fixed motion prototypes. Its benefit was concentrated at the single-step budget, whereas MAM alone performed best with two steps, indicating that a fixed prior can constrain generation when more refinement is affordable. A prior whose number of prototypes or strength adapts to the available computation could keep the single-step gain without this cost. Third, OpenFlyV2 was trained without real-world flight trajectories, and physical outcomes were judged by a human observer on a single platform. Whether training on physical flight data would further improve real-world performance remains open; answering it will require instrumented outcome measurement and evaluation beyond a single platform, including open agile research quadrotors~\cite{r37}. How much general vision-language capability is retained after aerial specialization also remains to be measured. Finally, the navigation baselines exceeded the memory of the onboard computer and were therefore compared with OpenFlyV2 in simulation only.

Language-guided inspection could combine navigation to a described landmark with questions about the resulting view. Because the perception--action loop does not depend on offboard computation, such inspection could extend to settings with limited connectivity, such as post-disaster sites. Extending this interaction to dialogue, reference tracking across requests, and concurrent navigation and question answering could support more flexible missions. Aerial vision-and-dialog navigation and online-dialogue action models provide related settings for studying clarification and interaction during flight~\cite{r38,r39}.

\section{Materials and Methods}

\subsection{Study design}

The objective of this study was to determine whether a single compact vision-language-action model could support aerial scene understanding and VLN while running entirely onboard a quadrotor. The study comprised four components. First, we constructed OpenFlyV2-130K, a simulation-based dataset of navigation trajectories and VQA pairs from 13 environments, three of which were excluded from all training. Second, we trained OpenFlyV2 in two supervised stages on both tasks, followed by reinforcement post-training of the navigation policy. Third, we evaluated one final checkpoint on five vision-language tasks (300 examples each) against three general-purpose vision-language models and on 1,100 simulated navigation episodes against four open-source navigation baselines, and we assessed the contributions of MAM and MSP through supervised ablations at one and two flow-integration steps. Fourth, we deployed the same checkpoint on a DJI M30T quadrotor with an onboard NVIDIA Jetson Orin NX and conducted 300 physical flight trials at 10 sites spanning university campuses, factories, parks, and urban areas, including 8 that were absent from the training data. All methods were evaluated on the same simulated episodes with shared success criteria, and physical-flight outcomes were judged by a human observer using the criteria described in Evaluation protocols.

\subsection{Dataset generation and annotation}

The environments summarized in Fig.~2 supplied rendered observations for both tasks in our simulation-based collection. AirSim~\cite{r23} and UE scenes provided explicit geometry and configurable camera viewpoints. The campus and post-disaster scenes were reconstructed using 3DGS~\cite{r24}. Eleven of the 13 environments, all except the two post-disaster ruins, were adapted from OpenFly~\cite{r11}; all trajectories, instructions, and VQA pairs were newly generated for this study.

Landmarks, obtained from ground-truth annotations and from manual annotation of the 3DGS scenes, were stored in a grid map. Each landmark record contained a semantic label, a 3D position, dimensions, and a local point cloud. To generate a trajectory, we randomly selected a target landmark and sampled an unoccluded starting viewpoint based on projections of nearby scene assets. This process was repeated to generate a continuous multi-landmark trajectory. Finally, the trajectory was optimized, and the corresponding images and camera poses were collected.

GPT-5~\cite{r40} generated navigation instructions from observation sequences and question--answer pairs from individual rendered aerial views. The VQA training subset was sampled from the generated annotation pool. Navigation evaluation episodes were sampled from a held-out test subset of the collection.

All generated annotations underwent human verification by a pool of 200 annotators. For navigation instructions, reviewers checked that landmark descriptions allowed accurate and unambiguous localization of the intended targets. For question--answer pairs, reviewers assessed the clarity of object references and the correctness of answers relative to the questions and observations. Of the generated samples, 60\% passed verification and were retained in the final dataset.

\subsection{OpenFlyV2 architecture}

OpenFlyV2 combines a shared Qwen3.5-0.8B backbone with a compact action expert (Fig.~6A). At navigation decision step $t$, the model receives the current aerial image $I_t$, a language instruction, the robot state $s_t$, and MAM tokens encoding recent visual and action history. It predicts an action chunk $\hat{A}_t \in \mathbb{R}^{H \times 4}$ with horizon $H=12$. Each action, corresponding to one waypoint, contains three body-frame motion increments and a stop signal. The XYZ components are normalized using min--max statistics from the training data. For vision-language queries, the same backbone instead decodes textual responses grounded in the current view. The output pathway is selected according to the task.

\begin{figure*}[t]
\centering
\includegraphics[width=\textwidth]{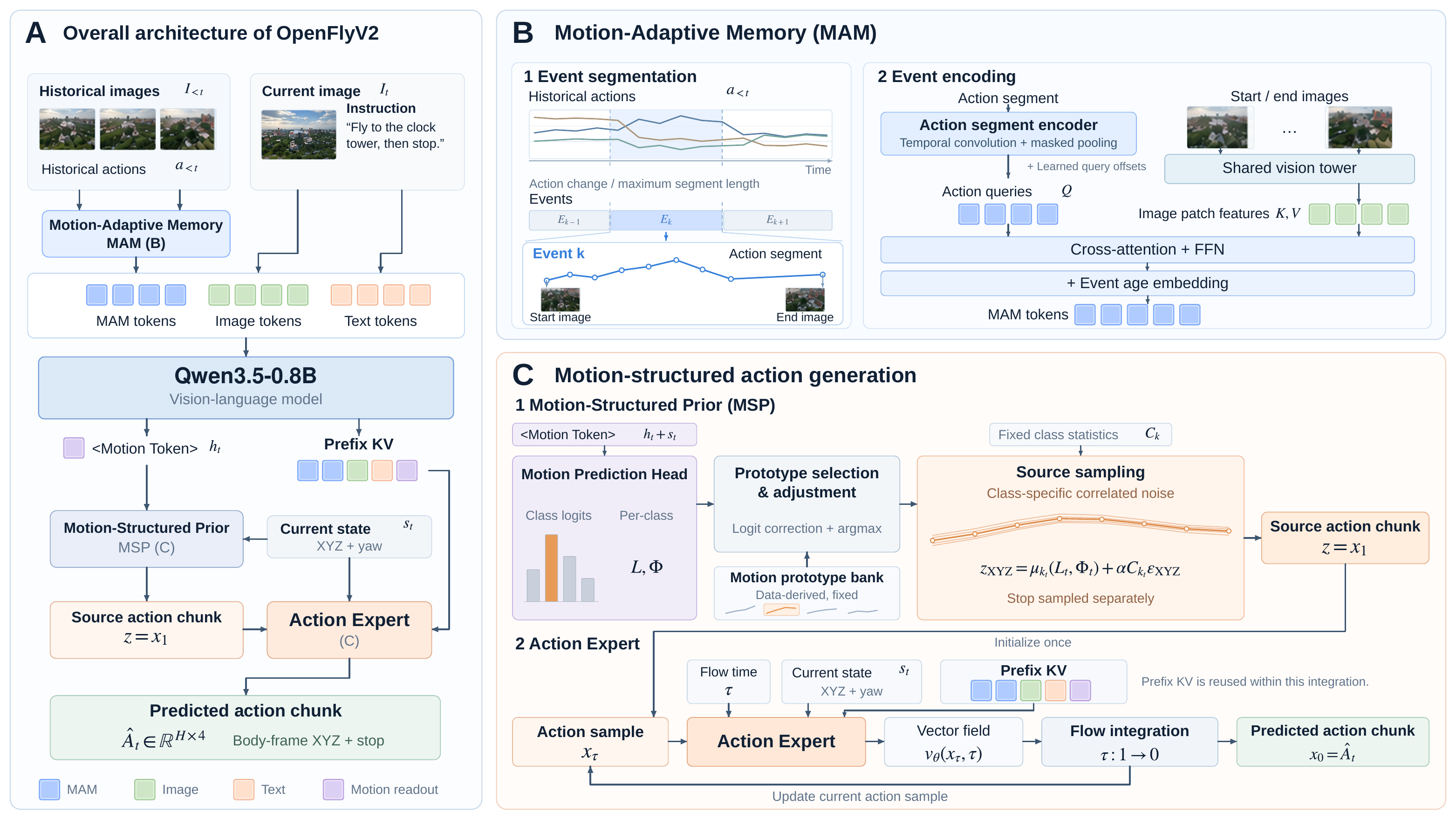}
\caption{\textbf{Architecture of OpenFlyV2.} \textbf{(A)} A shared vision-language backbone encodes the current image, instruction, and memory tokens to condition action generation. \textbf{(B)} Motion-Adaptive Memory (MAM) compresses historical action segments and their visual anchors into compact event-level tokens. \textbf{(C)} The Motion-Structured Prior (MSP) selects and adjusts a data-derived motion prototype to construct the action source distribution. The action expert transforms a sample from this distribution into the final action chunk through conditional flow matching.}
\label{fig:6}
\end{figure*}

Body-frame increments provide a local motion representation as the UAV changes position and heading. Action chunking predicts several successive actions within one output~\cite{r41,r42}, while subsequent prediction cycles incorporate new observations. Each flow-integration step updates the entire 12-waypoint chunk, so single-step generation still predicts all 12 entries.

The vision-language prefix comprises MAM tokens, current-image tokens, language tokens, and a designated motion token. The backbone provides two complementary conditions for action generation: the motion-token hidden state $h_t$ and the complete encoded prefix $c_t$. Together with $s_t$, $h_t$ conditions the MSP, which constructs an initial action sample. The action expert then transforms this sample into the final action chunk while attending to the full prefix. Its suffix comprises a projected state token and action tokens encoding the current action sample and flow time.

At the start of each action-generation cycle, the prefix is encoded using the latest observation and updated memory. Its cache is reused across the flow-integration steps for that chunk, avoiding repeated processing of unchanged visual and language tokens. Only the expert suffix is recomputed during flow integration; the prefix is refreshed for the next prediction cycle.

\subsection{Motion-adaptive memory}

MAM represents recent flight history as a bounded collection of motion events (Fig.~6B). Historical actions are processed causally: an event boundary is proposed when the change between consecutive actions exceeds a threshold or the segment reaches a prescribed maximum length, subject to a minimum separation between events. Each event contains the corresponding action segment and images anchored at its start and end.

For event $j$, a temporal convolutional encoder and masked pooling map the action segment to a compact representation. Projection and learned token offsets produce $P$ distinct action queries $Q_j$. The two anchor images are processed by the vision tower shared with the current image, yielding visual patch features $V_j$. Cross-attention uses the action queries to retrieve information from these visual features, followed by residual feed-forward processing and output normalization. Event memory is formed as
\begin{equation}
    M_{t,j}
    =
    F_{\mathrm{MAM}}(Q_j,V_j)
    +
    E_{\mathrm{age}}(d_{t,j}),
    \label{eq:mam_event}
\end{equation}
where $F_{\mathrm{MAM}}$ denotes the action-conditioned visual encoder and $E_{\mathrm{age}}$ encodes event age $d_{t,j}$. Up to $M$ events contribute at most $MP$ memory tokens to the vision-language prefix, with padding entries masked. Adding age information after visual-action fusion allows completed-event content to be cached separately from its changing age. This representation replaces dense historical image sequences with compact, motion-conditioned visual context. The learned offsets allow action queries to retrieve different aspects of an event, yielding $P$ tokens regardless of segment length. The limits $M$ and $P$ control the number of retained events and the capacity per event, respectively; the current image remains directly available outside this history budget. Completed-event content can be reused because its actions and anchor images no longer change; age is added when memory is assembled for a new prediction. This content cache can persist across prediction cycles, whereas the encoded-prefix cache is reused only within one cycle.

\subsection{Motion-structured action generation}

MSP introduces data-derived motion structure into the source distribution used for flow-based action generation. Offline, a subset of nonstationary training action chunks is reconstructed as three-dimensional trajectories relative to the chunk origin and normalized by total path length. The flattened trajectory shapes are clustered using $K$-means with $K=4$. The resulting centers define fixed motion prototypes containing stepwise horizontal-distance proportions, yaw profiles, and vertical profiles. The remaining training chunks are assigned to the nearest fixed center to obtain motion-class labels and within-class action statistics. Path-length normalization separates trajectory shape from travel distance, allowing each prototype to be adjusted to the extent and turning required by the current context.

A motion prediction head receives the contextual representation $h_t$ and current state $s_t$ (Fig.~6C). It predicts class logits and class-specific coefficients for motion length and accumulated yaw. The class $k_t$ with the highest prior-corrected logit is selected together with its predicted coefficients $L_t$ and $\Phi_t$. These coefficients adjust the selected prototype's horizontal extent, vertical profile, and accumulated yaw. The adjusted prototype is converted to body-frame XYZ motion and mapped to the normalized action space, yielding a mean trajectory $\mu_{k_t}(L_t,\Phi_t)$. The prediction head is supervised using nearest-center class labels and length and accumulated-yaw targets derived from training trajectories.

Let $\mu_{k_t} \in \mathbb{R}^{3H}$ denote the flattened XYZ mean and $C_{k_t}$ the Cholesky factor of the corresponding within-class action covariance. The source chunk is sampled as
\begin{equation}
    \begin{aligned}
        z_{\mathrm{XYZ}}
        &= \mu_{k_t}(L_t,\Phi_t)
           + \alpha C_{k_t}\epsilon_{\mathrm{XYZ}},\\
        z_{\mathrm{stop}} &= \epsilon_{\mathrm{stop}},\\[2pt]
        \epsilon_{\mathrm{XYZ}} &\sim \mathcal{N}(0,I_{3H}),\\
        \epsilon_{\mathrm{stop}} &\sim \mathcal{N}(0,I_H),
    \end{aligned}
    \label{eq:msp_source}
\end{equation}
where $\alpha$ controls the perturbation scale. The XYZ covariance captures correlations across action steps and coordinates; fixed coordinates are excluded from adjustment and perturbation. The XYZ and stop components are assembled into $z \in \mathbb{R}^{H \times 4}$. Thus, MSP selects and adjusts one motion prototype, and the action expert generates the final trajectory. The adjusted mean supplies a coherent motion pattern, while the covariance retains variation across coordinates and time. Stop entries begin as an independent Gaussian component and are refined jointly with XYZ by the expert.

Flow matching learns a transport velocity field, with conditional formulations supporting flexible source distributions~\cite{r43,r44}. The action expert transforms $z$ through conditional flow matching~\cite{r31} (Fig.~6C). Let $A_t^*$ denote a ground-truth action chunk and $\tau \in [0,1]$ the sampled flow time, distinct from the navigation step $t$. For action entries to be generated, training uses
\begin{equation}
    x_\tau
    =
    \tau z + (1-\tau)A_t^*,
    \qquad
    u_\tau
    =
    z-A_t^*.
    \label{eq:flow}
\end{equation}

Conditioned on the full prefix $c_t$ and state $s_t$, the expert predicts an action-space velocity field $v_\theta$. Its objective is
\begin{equation}
    \mathcal{L}_{\mathrm{FM}}
    =
    \mathbb{E}\left[
    \left\|
    v_\theta(x_\tau,\tau \mid c_t,s_t)-u_\tau
    \right\|_2^2
    \right],
    \label{eq:fm_loss}
\end{equation}
evaluated only on valid, nonconditioned action entries. Source construction uses predicted classes and coefficients during both training with MSP and deployment.

At inference, integration starts from $x_1=z$ and proceeds toward $\tau=0$ using Euler updates,
\begin{equation}
    \begin{aligned}
        x_{\tau+\Delta\tau}
        &= x_\tau+\Delta\tau\,
           v_\theta(x_\tau,\tau\mid c_t,s_t),\\
        &\hspace{1em}\Delta\tau<0.
    \end{aligned}
    \label{eq:flow_inference}
\end{equation}

The terminal sample $x_0$ is the predicted action chunk. Motion structure conditions the starting distribution without restricting the output to the selected prototype. The velocity field acts on candidate action chunks; only the terminal sample is converted to flight commands. One-step generation applies a single Euler update from $\tau=1$ to $\tau=0$; additional steps reevaluate the field at intermediate samples.

\subsection{Training}

Training comprised two supervised stages followed by reinforcement post-training of the navigation policy. We first trained the Base model jointly on the VLN and VQA training subsets. For navigation, the visual input consisted of four historical images together with the current image. Neither MAM nor MSP was enabled at this stage. VLN examples provided action supervision, while VQA examples provided question--answer supervision. This stage was intended to establish the model's ability to use historical observations before introducing a compressed history representation.

We then enabled MAM and MSP and continued joint supervised training. MAM replaced historical-image inputs with event-level tokens, while MSP supplied the motion-conditioned source distribution; the current image remained directly available. Both supervised stages optimized all parameters of their active modules. For Table~1, all variants underwent supervised training and were evaluated before reinforcement post-training. The Base model retained explicit visual history throughout.

Following supervised training, we post-trained the navigation policy by adapting group-relative optimization to a stochastic flow policy~\cite{r45,r46}. We sampled 4,000 tasks from training trajectories in 8 of the 10 training environments, excluding the two post-disaster ruins. Online rollouts began at recorded turning or stopping points, with preceding observations supplied by the expert trajectory and subsequent observations generated by the simulator. Perturbations to eligible turning starts exposed the policy to recovery decisions. Only the action expert was updated; the visual encoder, vision-language backbone, MAM, and MSP remained fixed.

Four rollouts sharing the same initial condition formed each group. Their returns combined task progress and terminal outcomes with heading and recovery terms, while penalizing collisions, invalid stopping, execution length, and positional deviation. Group-relative advantages supplied a trajectory-level learning signal, and regularization against the frozen initial expert constrained policy updates. Reinforcement post-training used two stochastic flow-integration transitions per action chunk. Final evaluation and physical deployment used one integration step without added transition noise, while retaining the sampled MSP source.

\subsection{Evaluation protocols}

Vision-language evaluation comprised 300 examples per task. Color recognition used multiple-choice answers scored against ground-truth annotations. Distance estimation and spatial relationship reasoning used free-form responses. Ground-truth distances were computed from point clouds as distances to the building surface. Distance estimates were assessed using mean absolute error and mean absolute relative error, calculated by dividing each absolute error by its ground-truth distance and averaging over parseable numerical responses. Visual grounding included 250 examples with visible targets and 50 with absent targets. Acc@0.5 was calculated over the 250 visible-target examples. GPT-6.1 Sol~\cite{r47} assessed spatial responses against reference answers and scored object descriptions on a 10-point scale. Evaluation details are provided in section S4; task-specific denominators, results, and 95\% confidence intervals are provided in table S8 (section S6). Per-question responses and scores are provided in data file S1.

Simulated navigation was evaluated on 800 episodes in seen environments and 300 in unseen environments, sampled from the held-out test subset. The baselines and OpenFlyV2 used the same evaluation episodes. Each baseline received inputs according to its original model requirements. Episodes ended when the model issued a stop command or after 250 execution steps, whichever occurred first. Success required a terminal goal distance below 25 m and no detected collision throughout the episode. The terminal goal distance was the minimum of the XY-plane distance to the reference trajectory endpoint, the XY-plane distance to the annotated target-building position, and, when valid, the nearest 3D distance to the target-building point-cloud surface. OSR measured whether a trajectory reached the goal region at any point. NE was the mean terminal goal distance defined above, and SPL combined successful completion with path efficiency, using the expert trajectory as the reference path. Main comparisons report metrics separately for seen and unseen environments. Architectural ablations pool both sets of episodes and compare checkpoints after supervised training, before reinforcement post-training, at one or two flow-integration steps. The final policy used one integration step. The numerical navigation comparison is summarized in table S9.

Physical evaluation comprised 300 trials, with 150 single-landmark and 150 two-landmark instructions, across 2 seen and 8 unseen locations. A human observer assessed success from an instruction-compliant flight path and the destination landmark appearing in the central or lower-central region of the final camera image. Two-landmark tasks required reaching both landmarks in the instructed order; skipping the first landmark counted as failure. No fixed trial time limit was imposed. When the observer identified a clear deviation from the instructed route, the flight was manually terminated and recorded as a failure. The overall SR pooled all 10 locations and both instruction types. The complete 300-trial register, including original instructions, video paths, and outcomes, is provided in section S7 and table S10.

\subsection{Onboard deployment}

For physical execution, predicted motion increments were denormalized and used by an onboard path tracker to generate body-frame velocity and yaw-rate commands for the DJI flight controller. For safety, flight altitude was held constant during physical trials. Waypoint progression depended on measured position rather than a fixed time interval. Position and heading came from DJI flight-controller estimates and were expressed relative to the task starting position and heading. Section S5 provides the state sources and control settings.

Real-time chunking addresses the delay between generating and executing successive action chunks~\cite{r48}. To maintain command availability during inference, we used asynchronous chunk execution with training-time action-prefix conditioning~\cite{r49}. The next chunk was generated while the current chunk was still executing, conditioned on commands committed during the inference interval. Training loss was applied to the remaining entries, teaching the expert to continue the committed motion without inference-time gradient guidance. Memory describes observations and motion already experienced, whereas the committed prefix supplies the boundary condition for continuing execution; the latest observation and updated memory inform the uncommitted portion of each new chunk.

\subsection{Statistical analysis}

SR and OSR were calculated as proportions of evaluated episodes or trials. SPL was averaged across all episodes, with failed episodes contributing zero, and NE was the mean terminal goal distance. The vision-language metrics were computed as described in Evaluation protocols. Onboard latencies were reported as the arithmetic mean of five timed runs after five warm-up runs. Tabulated benchmark results, per-episode navigation outcomes, per-question vision-language results, real-world trial records, and other reported quantitative aggregates are provided in data file S1.

For Fig. 4B, two-sided 95\% confidence intervals were estimated using 10,000 environment-stratified bootstrap resamples. Within each seen or unseen set, 100 episodes were sampled with replacement from each environment, using the same episode indices across models. Each metric was recomputed across all sampled episodes, and intervals were defined by the 2.5th and 97.5th percentiles. These intervals quantify episode-sampling uncertainty conditional on the evaluated environments and fixed model checkpoints. For the physical-flight SRs and for SR and OSR in Table 1, 95\% confidence intervals were computed with the Wilson score method, treating trials or episodes as independent. For the vision-language metrics in Fig. 4A, 95\% percentile confidence intervals were estimated from 10,000 bootstrap resamples of the evaluated examples, drawn separately for each model and task, and are reported in table S8.

Comparisons between OpenFlyV2 and each other model in Fig. 4 were paired by episode or question. Binary outcomes (SR, OSR, color recognition, spatial relationship reasoning, and Acc@0.5) were compared using two-sided exact McNemar tests. Continuous outcomes (SPL, NE, description score, and distance-estimation errors) were compared using two-sided paired sign-flip permutation tests on the mean difference with 9,999 resamples; distance-estimation comparisons used the examples with parseable responses from both models. Within each metric and evaluation set, \textit{P} values were adjusted for comparisons against multiple models using the Holm method, and adjusted \textit{P} < 0.05 was considered significant. All \textit{P} values are reported in table S11 and data file S1.

\end{document}